\documentclass[11pt,a4paper]{article}
\usepackage[hyperref]{emnlp-ijcnlp-2019}
\usepackage{times}
\usepackage{latexsym}

\usepackage{tabularx} % extra features for tabular environment
\usepackage{amsmath,amssymb}  % improve math presentation
\usepackage{graphicx} % takes care of graphic including machinery
\usepackage{cite} % takes care of citations
\usepackage{url}
\usepackage{xspace}
\usepackage{algorithmicx,algorithm}
\usepackage{algpseudocode}
\usepackage{mathtools}
\usepackage[inline]{enumitem}
\usepackage{booktabs}
\usepackage{bbm}
\usepackage{colortbl}
\usepackage{stmaryrd}
\usepackage{xcolor}
\usepackage{xcolor}
\usepackage{caption}
\usepackage{subcaption}
\usepackage{arydshln}
\usepackage{pifont}
\usepackage{multirow}
\usepackage{todonotes}
\usepackage{amsthm}
\usepackage{etoolbox}

\AtBeginEnvironment{quote}{\small}

\definecolor{darkgreen}{rgb}{0.0, 0.5, 0.0}

\usepackage{url}

\aclfinalcopy % Uncomment this line for the final submission

\newcommand{\first}{\textsc{First50}\xspace}
\newcommand{\perfecttrans}{\textsc{PerfectTrans}\xspace}
\newcommand{\transthensum}{\textsc{TransThenSum}\xspace}
\newcommand{\bleu}{\textsc{Bleu}\xspace}
\newcommand{\rougeone}{\textsc{Rouge-1}\xspace}
\newcommand{\rougetwo}{\textsc{Rouge-2}\xspace}
\newcommand{\rougel}{\textsc{Rouge-l}\xspace}
\newcommand{\rouge}{\textsc{Rouge}\xspace}
\newcommand{\urls}[1]{{\scriptsize \url{#1}}}
\renewcommand{\texttt}[1]{{\fontfamily{lmtt}\selectfont #1}}

\newcommand{\gvsnippet}{\texttt{gv-snippet}\xspace}
\newcommand{\gvcrowd}{\texttt{gv-crowd}\xspace}

\newcommand{\idcs}{${}^\odot$}

\newcommand{\idmsr}{${}^\heartsuit$}

\title{Global Voices: Crossing Borders in Automatic News Summarization}

\author{Khanh Nguyen\idcs \and Hal Daum{\'e} III\idcs\idmsr \\
  University of Maryland, College Park\idcs, Microsoft Research, New York\idmsr\\
  \tt{ kxnguyen@cs.umd.edu \ \ hal@umiacs.umd.edu }
 }

\date{}

\begin{document}

\maketitle
\begin{abstract}
We construct \emph{Global Voices}, a multilingual dataset for evaluating cross-lingual summarization methods. 
We extract social-network descriptions of Global Voices news articles to cheaply collect evaluation data for into-English and from-English summarization in 15 languages.
Especially, for the into-English summarization task, we crowd-source a high-quality evaluation dataset based on  guidelines that emphasize accuracy, coverage, and understandability.
To ensure the quality of this dataset, we collect human ratings to filter out bad summaries, and conduct a survey on humans, which shows that the remaining summaries are preferred over the social-network summaries. 
We study the effect of translation quality in cross-lingual summarization, comparing a translate-then-summarize approach with several baselines. 
Our results highlight the limitations of the \rouge metric that are overlooked in monolingual summarization. 
Our dataset is available for download at \url{https://forms.gle/gpkJDT6RJWHM1Ztz9} .
\end{abstract}

\section{Introduction}
Cross-lingual summarization is an important but highly unexplored task. The ability to summarize information written or spoken in any language at a large scale would empower humans with much more knowledge about the diverse world. 
Despite the fast development of automatic summarization \citep{allahyari2017text,dong2018survey,gambhir2017recent}, present technology mostly focuses on monolingual summarization. 
There are currently very limited standard, high-quality multilingual data for evaluating cross-lingual summarization methods (e.g., \citet{giannakopoulos-2013-multi,li-etal-2013-multi,elhadad-etal-2013-multi}). 
Two main challenges present in constructing such a dataset. First, the cost of crow-sourcing human-written summaries is high. It generally takes a long time for a human to summarize a document, as they not only have to read and understand information in the article, but also have to make complex decisions in sieving and paraphrasing the information. 
Second, it is difficult to design summarization guidelines for humans, as the task is generally not well-defined: the selection of what content is ``important" in a summary is based on subjective and common-sense rules that vary among individuals and are difficult to be expressed precisely in words.

\begin{figure}[t!]
    %\centering
    \includegraphics[width=0.9\linewidth]{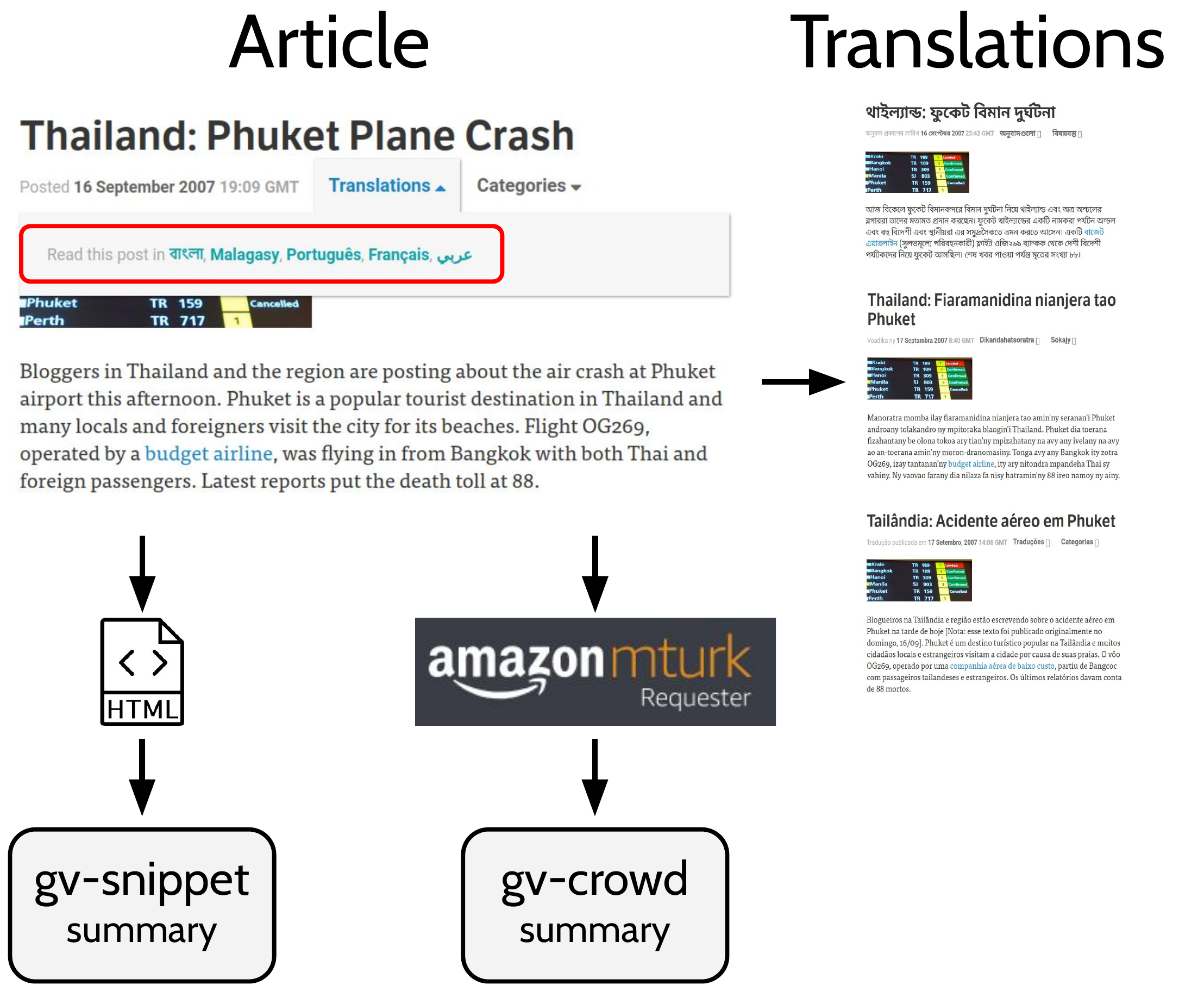}
    \caption{Data construction pipeline. We collect two types of summary: (a) the social network description of the article (\gvsnippet) and (b) the 50-word summary written by Mechanical Turk workers following our guidelines (\gvcrowd).   }
    \label{fig:sample}
\end{figure}

Even in monolingual summarization, there were only few attempts in constructing summarization datasets via crowd-sourcing \citep{over2007duc, dang2008overview,dang2010overview}.
These datasets are mostly used for evaluation due to their small sizes. 
To construct large-scale training datasets, researchers mine news sources that naturally provide human-written summaries \citep{hermann2015teaching, sandhaus2008nyt}, or construct artificial summaries from document titles \citep{rush2015neural}. 
Summaries collected in this way may be not best for evaluation because they are generated under unknown guidelines (or there may be no guidelines at all). 
Previous work on cross-lingual summarization performs evaluation with human judgments \citep{orǎsan2008evaluation}, or with automatic metrics and noisy source articles generated by automatic translation systems \citep{wan2010cross, ouyang2019robust}. The former approach is expensive and not reproducible, while the latter is prone to biases induced by translation systems that could be further amplified by summarization systems.

This paper presents \emph{Global Voices}, a high-quality multilingual dataset of summaries of news articles. 
The dataset can serve as a standard benchmark in both multilingual and cross-lingual summarization.
Global Voices\footnote{\urls{https://globalvoices.org/}} is a multilingual website that reports and translates news about unheard voices across the globe. 
Translation in this website is performed by the Lingua team,\footnote{\urls{https://globalvoices.org/lingua/}} consisting of volunteer translators.
As of August 2019, Global Voices provides translations of news articles in 51 languages; many articles are translated into multiple languages.
Figure \ref{fig:sample} illustrates a sample article from Global Voices. 
We extract the social-network descriptions of the articles to (cheaply) construct \gvsnippet, an evaluation set for multilingual and cross-lingual news summarization.
Nevertheless, these descriptions usually have poor coverage over the original contents because they were written with the intention of drawing user clicks to read more about the articles. 
Therefore, besides \gvsnippet, we construct a smaller but higher-quality dataset of human-written English summaries, called \gvcrowd, based on our guidelines which explicitly emphasize accuracy, coverage and understandability.
The Global Voices dataset is summarized in Table \ref{tab:dataset}.
It currently supports 15 languages, which span nine language genera
(Romance, Barito, Indic, Slavic, Semitic, Greek, Germanic, Japanese, Bantoid) and five language families (Indo-European, Austronesian, Japanese,
Niger-Congo, Afro-Asiatic).

\section{Dataset Construction}
\begin{table}[t!]
    \small
    \centering
    \setlength{\tabcolsep}{1.2pt}
	\begin{tabular}{lccc}
		\toprule
	     Language & ISO 639-1 & gv-snippet & gv-crowd \\
	     \midrule 
	     \multicolumn{4}{l}{\textbf{Number of articles}}\\
	     English & en & 4,573 & 529 \\
	     Spanish & es & 3,921 & 487 \\
	     Malagasy & mg & 2,680 & 374 \\
	     Bengali & bn & 2,253 & 352 \\
	     French & fr & 2,130 & 352  \\
	     Portuguese & pt  & ~~~798 & 162 \\
	     Russian & ru & ~~~795 & 139 \\
	     Arabic & ar & ~~~745 & 191 \\
	     Italian & it & ~~~718 & 135 \\
	     Macedonian & mk & ~~~701 & 138 \\
	     Greek & el & ~~~694 & 128 \\
	     German & de & ~~~647 & 204 \\
	     Japanese & ja & ~~~424 & ~~75 \\
	     Swahili & sw & ~~~418 & ~~84 \\
	     Dutch & nl & ~~~348 & ~~87 \\
	     \midrule 
	     \multicolumn{4}{l}{\textbf{Other statistics}} \\
	     \multicolumn{2}{l}{Summarized by} & GV authors  & MTurkers \\
	     \multicolumn{2}{l}{Summary languages} & All versions & English \\
	     \multicolumn{2}{l}{Summary lengths (words)} & - & 40-50 \\
	     \multicolumn{2}{l}{Article lengths (words)} & 150-500 & 150-350 \\
	     
        \bottomrule
	\end{tabular}
	\smallskip
	\caption{Summary of the Global Voices dataset. The dataset include articles in 15 languages. English versions of all non-English articles are included. 
	The \gvsnippet split contains social-network summaries of all articles, while the \gvcrowd split contains crowd-sourced summaries of English articles.  }
	\label{tab:dataset}
\end{table}

\noindent\textbf{Data Collection and Pre-Processing}. Using Scrapy,\footnote{\urls{https://scrapy.org/}} we crawl and download HTML source codes of 41,939 English articles and their translations.  We use \texttt{bs4}\footnote{\urls{https://www.crummy.com/software/BeautifulSoup/}} to extract each article's main content and remove image captions. Next, we use \texttt{html2text}\footnote{\urls{https://pypi.org/project/html2text/}} to convert the main content's HTML source code to regular text, removing web-page and image URLs. 
Since an article may content block-quotes written in original languages, we detect language of each paragraph and remove paragraphs that are not in the article's main language. 
Language detection is conducted by voting decisions of four packages: \texttt{langdetect},\footnote{\urls{https://pypi.org/project/langdetect/}} \texttt{langid},\footnote{\urls{https://github.com/saffsd/langid.py}} \texttt{polyglot},\footnote{\urls{https://pypi.org/project/polyglot/}} \texttt{fastText}\footnote{\urls{https://fasttext.cc/docs/en/language-identification.html}} \citep{joulin2016fasttext,joulin2016bag}. 

\noindent\textbf{Constructing \gvsnippet}. This split includes articles whose English versions contain from 150 to 500 words. For each article, we extract its Open Graph description by extracting the \texttt{meta} tag with property \texttt{og:description} in the HTML source code, and use the description as the reference summary of the article. These descriptions are short text snippets that serve as captions of the articles when they appear on social networks (e.g. Facebook, Twitter). 

\noindent\textbf{Crowd-Sourcing \gvcrowd}. We select English articles that contain 150-350 words, and request workers from Mechanical Turk\footnote{\urls{https://www.mturk.com/}} (MT) to summarize them in 40-50 words. Each HIT\footnote{a Mechanical Turk task.} asks a worker to summarize five articles in 35 minutes. We recruit Turkers in Canada and the U.S.A. with Masters qualification, a HIT approval rate greater than or equal to 97\%, and a number of HITs approved greater than or equal to 1,000. On average, collecting a summary costs 1.50 USD (including taxes and extra fees). We inform workers of our evaluation guidelines, which focus on three criteria:
\begin{itemize}
    \item \emph{Accuracy}: information in a summary should be based on the original article only. It can be paraphrased from but should not disagree with information in the article.
    \item \emph{Coverage}: a summary should reflect the most important messages/stories in the original article. Each message/story should be captured as detailed as possible, without missing other important messages/stories.
    \item \emph{Understandability}: a summary must be written in standard, fluent English. Readers must be able to understand the summary without reading the original article. Understanding the summary must not require any additional knowledge beyond knowledge required to understand the article. 
    
\end{itemize} In comparison, the DUC-2004 dataset \citep{over2007duc} only provides subtle format suggestions and leaves the summary contents almost entirely to the decisions of the writers:

\begin{quote}
 ``...Imagine that to save time, rather than read through a set of complete documents, you first read a list of very short summaries of those documents and based on these summaries
 you choose which documents to read in their entirety. Create your very short summaries to be useful in such a scenario. A very short summary could look like a newspaper headline, be a list of important terms or phrases separated by commas, a sentence, etc. It should not contain any formatting, i.e., no indented lists, etc. Feel free to use your own words."
 
 Source: {\scriptsize\url{https://duc.nist.gov/duc2004}}
\end{quote} Our guideline criteria are similar to those of the TAC 2010's guided summarization task \citep{dang2010overview} but we do not restrict the summary format using domain-specific templates. 

Some articles may read disrupted due to removals of images and videos, and may contain non-English texts. To ensure the summaries are based on the English texts only, we advise workers to (a) \emph{not} web-search for the original content and (b) ignore the non-English contents. We also emphasize spelling words correctly and recommend copying difficult-to-spell words from the original articles. 
In the end, we collect 840 summaries for 738 articles.

\noindent \textbf{Human Evaluation of \gvcrowd}. The summary-collecting task receives mostly positive feedback from workers. The task is widely regarded as ``fun", ``interesting", and ``challenging". However, many workers raised concern about the strict time constraint. 
To evaluate the quality of the dataset, we launch another MT task in which we ask workers to rate and post-edit the summaries collected in the previous task. 
Each task HIT requires evaluating ten summaries in 60 minutes.
We recruit workers in Canada and the U.S.A. with a HIT approval rate greater than or equal to 97\%, and a number of HITs approved greater than or equal to 1,000.

Specifically, we ask workers to provide two types of ratings: \textit{criterion-based} ratings and \textit{overall} ratings. 
Each worker is instructed to first give a 1-to-5 rating of a summary in each of our three criteria (accuracy, coverage, understandability), and then to give an overall rating of the summary.
We define three levels of the overall rating:
\begin{itemize}
    \item {\color{red}\emph{Bad}}: the summary misrepresents the original article. It contains factual errors that disagree with the content of the article. OR it does not cover the most important message/story of the article. OR it is missing other important points that could easily be included without violating the 50-word constraint. 
    \item {\color{blue}\emph{Acceptable}}: the summary covers the most important message/story of the article. It does \emph{not} contain factual errors. It is missing one or two important points that would be difficult to include in a 50-word summary.
    \item {\color{darkgreen}\emph{Good}}: the summary covers the most important message/story of the article. It does \emph{not} contain factual errors. All important points are captured.
\end{itemize} 
 In addition, the worker is required to write short reasons (each in 5-25 words) to justify their ratings.

Among 840 summaries collected, 383 (45.60\%) were rated as {\color{darkgreen}\emph{Good}}, 264 (31.43\%) {\color{blue}\emph{Acceptable}}, and 193 (22.98\%) {\color{red}\emph{Bad}}. 
We observe that among the three criteria, understandability is easiest to meet while coverage is the most challenging: the mean understandability rating is 4.06 while the mean coverage rating is only 3.47; about 90\% of the summaries attain understandability ratings of at least 3.
By computing Pearson correlation coefficients, we find that the overall rating most strongly linearly correlates with the coverage rating (0.81) and least with the understandability rating (0.57).
Common flaws identified by the human evaluators include: missing important points, factual errors, abstruse and/or verbose writing. 

To construct the \gvcrowd split, we pair each article with its highest-rated summaries\footnote{For a pair of summaries, we first compare their overall ratings, then sums of three criterion-based ratings, then the individual accuracy, coverage, understandability ratings (in this specific order).} and excluded articles that (a) are paired with {\color{red}\emph{Bad}} summaries or (b) have a criterion-based rating below 3. 
We also ask workers to correct spelling and factual errors in the {\color{red}\emph{Bad}} summaries, but these post-edited summaries require further evaluation to be included in the dataset in the future. 
To facilitate summarization evaluation studies, we will release all the summaries accompanied with their ratings, reasons, and post-edit versions. 

For a (randomly selected) subset of 50 articles, we collect \emph{three} summaries per article to study the diversity in quality and language usage among human-written summaries of the same documents. 
We find that the summary quality does not vary greatly: the overall-rating difference between the highest and lowest rated summaries is at most 1 in 74\% of these articles.  
To quantify the diversity of summaries, we calculate the \emph{pairwise} \rouge scores, using one summary as the reference and another as the predicted
\begin{align}
    \rouge_{\textrm{pair}} &= \frac{1}{3 \cdot 50}  \sum_{i = 1}^{50} \sum_{1 \leq j < k \leq 3} \rouge(s_{i, j}, s_{i, k}) 
\end{align} where $s_{i, j}$ and $s_{i, k}$ are distinct summaries of the $i$-th article. The \textsc{Rouge}$_{\textrm{pair}}$\textsc{-1,2,L} F-1 scores are relatively low (39.44, 12.39, and 32.85, respectively), indicating that the summaries highly vary in vocabulary and sentence structure.

\noindent\textbf{Human Comparison of \gvsnippet and \gvcrowd}. To ensure that the \gvcrowd summaries are of higher quality than the \gvsnippet summaries, we conduct a survey that asks MT workers to compare the two types of summary. Concretely, each worker reads an article and its \gvsnippet and \gvcrowd summaries. We ask the worker to specify which summary (or none) is better in each of the three criteria and is better overall. 
We remove partial sentences that end with ``..." in the \gvsnippet summaries to ensure that the workers rate the two types of summary mainly based on their contents, not based on any peculiar features.
We also randomly shuffle the order of the summaries in a pair so that the workers cannot rely on the order to determine the summary type. 
Each worker is given 45 minutes to compare five summary pairs. 
Each summary pair is evaluated by three workers. 
We recruit workers with similar qualifications to those in the \gvcrowd evaluation task.

The outcome of this survey is positive. 
In 22 out of the 30 articles included in the survey (75.9\%), at least 2 out of 3 workers prefer the \gvcrowd summary.
Overall, 63 out of 90 workers (70.0\%) prefer the \gvcrowd summaries to its \gvsnippet counterparts.
As expected, coverage is the criterion where the \gvcrowd summaries show most strength against the \gvcrowd summaries, with a preference ratio of 83.3\% (25/30) compared to 66.7\% (20/30) of accuracy or understandability.  

\begin{figure}
    \centering
    \includegraphics[width=\linewidth]{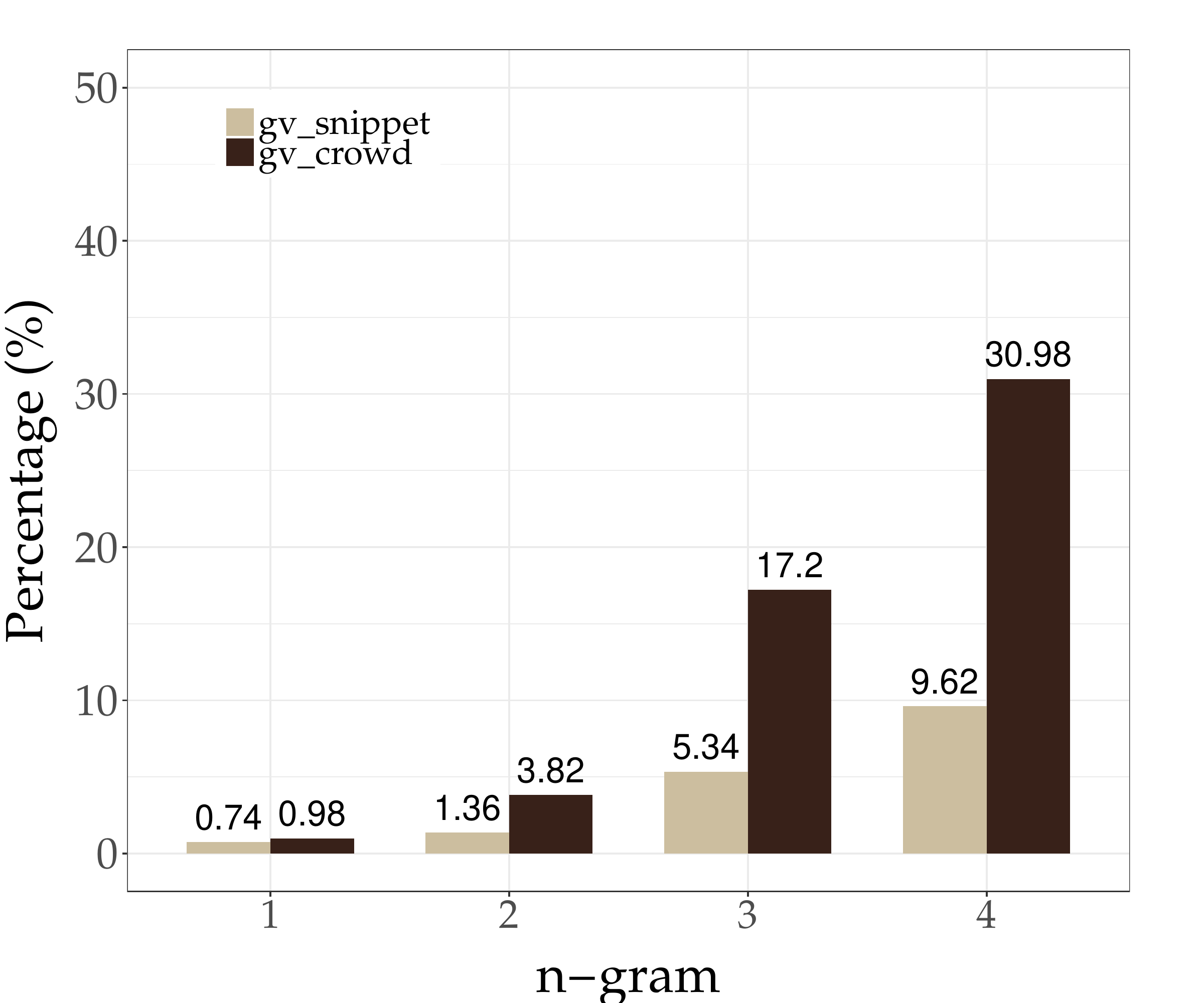}
    \caption{Average fraction of $n$-grams in the summary that are not seen in the original article.}
    \label{tab:novel_ngram}
\end{figure}

We also evaluate these two types of summary in terms of how novel their summaries are compared to the original articles. \autoref{tab:novel_ngram} shows the average fractions of novel $n$-grams of each type of summary. Overall, the summaries reuse most words in the articles. 
The \gvcrowd summaries contain substantially more novel 3-grams and 4-grams than the \gvsnippet summaries, partly because each sentence of a \gvcrowd summary usually includes information from multiple sentences in the original article. On 73\% of the articles in the \gvcrowd split, the \gvcrowd summary has higher fractions of novel $n$-grams than the \gvsnippet counterpart (with $n = 1,2,3,4$).

\begin{table}[t!]
    \small
    \centering
	\begin{tabular}{lcc}
		\toprule
	     Model & Train & Validation \\
	     \midrule
	     \multicolumn{3}{l}{\textbf{Translation} (sentences)} \\
	      \addlinespace[2.5pt]
	     Spanish-English & 4.1M & 3K \\
	     French-English & 5.6M & 3K \\
	     German-English & 151.6K & 2K \\
	     Arabic-English & 174.3K & 2K \\
	     \midrule 
	     \multicolumn{3}{l}{\textbf{Summarization} (pairs of documents and summaries)} \\
	      \addlinespace[2.5pt]
	     English & 287.2K & 13.4K \\
        \bottomrule
	\end{tabular}
	\smallskip
	\caption{Data used to train and validate translation and summarization models.}
	\label{tab:dataset}
\end{table}

\begin{figure*}
    \centering
    \includegraphics[width=\linewidth]{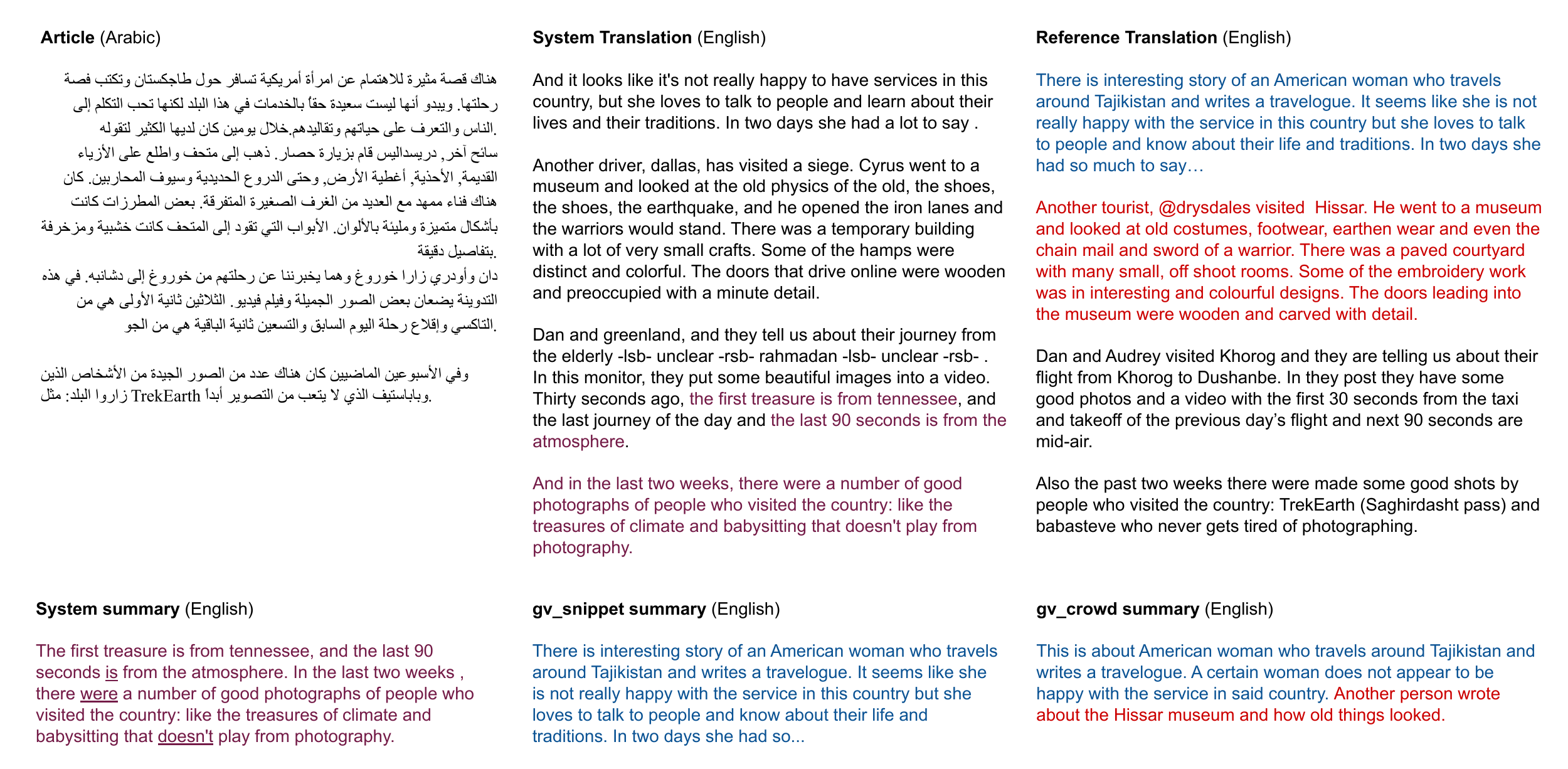}
    \caption{An example in our dataset. The source document is originally written in English and is translated into Arabic by a Global Voices translator. Our translation system translates the Arabic article into English poorly. The summarization system mostly copies segments from the translation and carries grammatical errors (underlined) from the translation to its summary. The \gvsnippet summary is a mere copy of the first few sentences of the English version of the article (though this may not always be the case in other articles). On the other hand, the \gvcrowd summary offers better coverage, including information in the second paragraph. Note that this article is challenging to summarize perfectly in 50 words because it features four different parallel stories at the same time. Here, the \gvcrowd summarizer trades off coverage for specificity of the stories. }
    \label{fig:example}
\end{figure*}
\section{Experiments}
\begin{table*}[t!]
    \small
    \centering
    \setlength{\tabcolsep}{5pt}
	\begin{tabular}{lcccc}
		\toprule 
		Method & Spanish-English & French-English & German-English & Arabic-English \\
	    \midrule
	    \multicolumn{5}{l}{\textbf{Translation quality} (\bleu $\uparrow$)} \\
	    \addlinespace[2.5pt]
	    Transformer & 37.45 & 29.80 & 19.34 & 10.77 \\
	    \midrule 
	    \multicolumn{5}{l}{\textbf{Summarization quality} evaluated on \gvsnippet (\textsc{Rouge-1$\mid$2$\mid$L} F-1 scores $\uparrow$)} \\
	    \addlinespace[2.5pt]
	    \first & 63.7 $\mid$ 55.1 $\mid$ 61.3 & 64.7 $\mid$ 56.2 $\mid$ 62.3 & 65.2 $\mid$ 57.1 $\mid$ 63.0 & 62.9 $\mid$ 53.5 $\mid$ 60.5 \\
	    \perfecttrans & 38.0 $\mid$ 22.1 $\mid$ 34.0 & 38.1 $\mid$ 21.8 $\mid$ 34.0 & 37.7 $\mid$ 21.9 $\mid$ 33.6 & 36.8 $\mid$ 20.0 $\mid$ 32.7 \\
	    \transthensum & 33.0 $\mid$ 12.4 $\mid$ 28.4 & 32.0 $\mid$ 10.6 $\mid$ 27.2 & 28.3 $\mid$ \xspace\xspace 7.4 $\mid$ 23.7 & 24.5 $\mid$ \xspace\xspace 4.3 $\mid$ 20.4  \\
	    \midrule
	    \multicolumn{5}{l}{\textbf{Summarization quality} evaluated on \gvcrowd (\textsc{Rouge-1$\mid$2$\mid$L} F-1 scores $\uparrow$)} \\
	    \addlinespace[2.5pt]
	    \first & 46.4 $\mid$ 23.4 $\mid$ 40.4 & 46.0 $\mid$ 22.8 $\mid$ 40.1 & 47.4 $\mid$ 25.7 $\mid$ 40.9 & 45.9 $\mid$ 22.9 $\mid$ 40.4 \\
	    \perfecttrans & 36.1 $\mid$ 13.5 $\mid$ 31.3 & 36.7 $\mid$ 13.7 $\mid$ 31.7 & 36.6 $\mid$ 14.1 $\mid$ 31.6 & 36.9 $\mid$ 14.0 $\mid$ 31.9  \\
	    \transthensum & 35.1 $\mid$ 10.6 $\mid$ 30.0 & 33.3 $\mid$ \xspace\xspace 8.9 $\mid$ 28.5 & 29.4 $\mid$ \xspace\xspace 6.0 $\mid$ 25.0 & 26.0 $\mid$ \xspace\xspace 3.8 $\mid$ 22.1 \\
        \bottomrule
	\end{tabular}
	\smallskip
	\caption{Cross-lingual summarization results with different approaches. Translation quality is measured on the \gvsnippet articles, of which the \gvcrowd articles are a subset.}
	\label{tab:results}
\end{table*}

We study the task of generating English summaries of non-English news articles.
This task can naturally be decomposed into two subtasks: translation and summarization. 
We follow a \emph{translate-then-summarize} approach where each article is first translated into English using a pre-trained machine translation model, then the translation is summarized using a pre-trained English summarization model. 
Data for training models in both subtasks are publicly available, allowing solving the joint task in a \emph{zero-shot} manner, in the sense that no parallel pairs of (original document, English summary) are provided during training.
On the other hand, a \emph{summarize-then-translate} approach is practically difficult to implement because of the lack of large-scale datasets for training reliable summarization models in  non-English languages. 

\noindent\textbf{Translation Models}. Our goal is to study the effect of translation quality in this task. Hence, we employ translation models trained under various amounts of resources.
We conduct experiments in four source languages: Spanish (es), French (fr), German (de), and Arabic (ar). 
Concretely, we train the \{es,fr\}-en models using the large-scale CommonCrawl and News Commentary datasets, and train the \{de,ar\}-en models using the low-resource multilingual TED \citep{duh18multitarget} dataset. We apply standard machine translation pre-processing steps, normalizing and tokenizing the data with Moses scripts. 
We tokenize Arabic texts with the PyArabic tool \citep{zerrouki2012pyarabic}. 
Our translation models implement the Transformer architecture \citep{vaswani2017attention}.
The \{es,fr\}-en models have the same hyperparameters as those of the base Transformer architecture described in Table 3 of \citet{vaswani2017attention}. 
The \{de,ar\}-en models have less parameters, using 4 attention heads and a feed-forward hidden size of 1024. We train the models using the \texttt{fairseq-py} toolkit \citep{ott2019fairseq}. 
Since the models are trained to perform sentence-level translation, we split the source articles into sentences, perform translation, and join the output sentences into articles. 
The training settings are the same as those of \citet{vaswani2017attention} except that: (a) the maximum tokens in a batch is 4,000, (b) the \{es,fr\}-en models and the \{de,ar\}-en models are trained for $5 \cdot 10^4$ and $8 \cdot 10^5$ iterations, respectively, and (c) the \{de,ar\}-en models use a dropout ratio of 0.3.
Training with an Nvidia Titan Xp GPU took place in approximately 5 hours for the smaller models and 3.5 days for the larger models.

\noindent \textbf{Summarization Models}. We employ the state-of-the-art Bi-LSTM bottom-up abstractive summarization model \citep{gehrmann2018bottom}. 
We make use of a pre-trained instance of this model provided by OpenNMT-py \citep{klein2017opennmt} and trained on the CNN/DailyMail dataset \citep{hermann2015teaching}. 

\noindent \textbf{Baselines}. We compare the following approaches:
\begin{itemize}[nolistsep]
    \item \first: copies the first 50 words of the English version of the source article. 
    \item \perfecttrans: directly summarizes the English version of the source article.
    \item \transthensum: our approach which first translates the source article into English then summarizes the translation. 
\end{itemize}

\noindent \textbf{Evaluation}. Translation quality is measured by corpus-level \bleu, treating each article as a data point. Summarization quality is determined by computing \rougeone, \rougetwo, \rougel F-1 scores.

\noindent \textbf{Results}. Table \ref{tab:results} presents our results. A qualitative example is illustrated in \autoref{fig:example}. As expected, translation quality varies among different pairs of languages. The Spanish-English model achieves the highest \bleu score (34.45) due to the amount of training data and the closeness between the language pair; on the other spectrum, the Arabic-English model offers poorest translations (10.77). 
Nevertheless, despite the large gaps in \bleu scores, we observe much smaller divergences in \rougeone and \rougel scores. 
For example, in the extreme case of Arabic-English, even when the \bleu drops by almost 90\% when switch from the reference to the predicted translations, the \rougel F1-score only decreases by only about 30\%.
This observation highlights a major limitation of \rougeone and \rougel: their insensitivity to the summary readability. Even though a source document may contain meaningless, ungrammatical contents (reflected by a low \bleu score), a model that summarizes by simply copying phrases can easily achieve high \rougeone and \rougel scores. 
This limitation is difficult to observe in the context of monolingual summarization because the source documents come from natural sources and thus are mostly grammatical and meaningful.
Another interesting finding is that the \first baseline achieves higher \rouge scores when evaluated on \gvsnippet than on \gvcrowd.
This observation indicates that the \gvsnippet summaries overlap highly with the beginning part of the articles, confirming the results from our human preference survey that these summaries generally have poorer coverage over the entire articles than the \gvcrowd summaries.

\section{Conclusion}
This work introduces a dataset for evaluating multilingual and cross-lingual summarization methods in multiple languages.  
Future work aims to extend the dataset to more languages and construct a large-scale training dataset. 
Another interesting direction is to study whether multi-task learning can benefit cross-lingual summarization. 
To take advantage of the fact that translating the entire source article may not be necessary, it would be useful to teach models to devise more efficient translation strategies by informing them of the downstream summarization objective.  
\section*{Acknowledgement}

This research is partially based upon work supported in part by the Office of the Director of National Intelligence (ODNI), Intelligence Advanced Research Projects Activity (IARPA), via contract \#FA8650-17-C-9117. The views and conclusions contained herein are those of the authors and should not be interpreted as necessarily representing the official policies, either expressed or implied, of ODNI, IARPA, or the U.S. Government. The U.S. Government is authorized to reproduce and distribute reprints for governmental purposes notwithstanding any copyright annotation therein.

This research is also partially based upon work supported by the National Science Foundation under Grant No. IIS-1618193. Any opinions, findings, and conclusions or recommendations expressed in this material are those of the author(s) and do not necessarily reflect the views of the National Science Foundation.

\bibliography{journal-full,emnlp-ijcnlp-2019}
\bibliographystyle{acl_natbib}

\end{document}